\documentclass[letterpaper]{article} 
\usepackage{aaai25}  
\usepackage{times}  
\usepackage{helvet}  
\usepackage{courier}  
\usepackage[hyphens]{url}  
\usepackage{graphicx} 
\usepackage{natbib}  
\usepackage{caption} 
\usepackage{algorithm}
\usepackage{algorithmic}

\usepackage{amssymb}
\usepackage{amsmath}
\usepackage{subfigure}

\usepackage{bibentry}
\usepackage{bbding}
\usepackage{longtable,booktabs,array}
\usepackage{newfloat}
\usepackage{listings}
\DeclareCaptionStyle{ruled}{labelfont=normalfont,labelsep=colon,strut=off} 
\floatstyle{ruled}
\newfloat{listing}{tb}{lst}{}
\floatname{listing}{Listing}
\title{Semi-IIN: Semi-supervised Intra-inter modal Interaction Learning Network for Multimodal Sentiment Analysis}
\author{
    Jinhao Lin,
    Yifei Wang,
    Yanwu Xu,
    Qi Liu\thanks{Corresponding author}
}
\affiliations{

    South China University of Technology\\
    {ftlinjinhao}@mail.scut.edu.cn, ywang634@Outlook.com, {ywxu}@ieee.org, {drliuqi}@scut.edu.cn\\
}

\usepackage{bibentry}

\begin{document}

\maketitle
\begin{abstract}
Despite multimodal sentiment analysis being a fertile research ground that merits further investigation, current approaches take up high annotation cost and suffer from label ambiguity, non-amicable to high-quality labeled data acquisition. Furthermore, choosing the right interactions is essential because the significance of intra- or inter-modal interactions can differ among various samples. To this end, we propose Semi-IIN, a Semi-supervised Intra-inter modal Interaction learning Network for multimodal sentiment analysis. Semi-IIN integrates masked attention and gating mechanisms, enabling effective dynamic selection after independently capturing intra- and inter-modal interactive information. Combined with the self-training approach, Semi-IIN fully utilizes the knowledge learned from unlabeled data. Experimental results on two public datasets, MOSI and MOSEI, demonstrate the effectiveness of Semi-IIN, establishing a new state-of-the-art on several metrics. Code is available at https://github.com/flow-ljh/Semi-IIN.
\end{abstract}

%
\section{Introduction}
Multimodal sentiment analysis (MSA) has attracted increasing attention in recent years due to the rapid development of online social media platforms \cite{poria2020beneath}. Multimodal data offers more emotional cues than unimodal sentiment analysis, allowing machines to interpret human behaviors better and make more precise sentiment predictions \cite{zhang2021deep,hu2021bidirectional}. However, utilizing various modalities for analyzing human emotions continues to be a significant obstacle, particularly in the context of multimodal interactions with unlabeled data. Existing methods can be categorized into supervised learning and semi-supervised learning approaches. The former focuses on multimodal fusion and alignment, where the goal is to extract complementary information from different modalities and better understand human emotions. For multimodal fusion, current approaches acquire joint representations by imposing constraints \cite{hazarika2020misa} or employing interactive operations \cite{zadeh2017tensor,liu2018efficient} on the representations of individual modalities within the feature space. For multimodal alignment, researchers are committed to designing a cross-modal attention mechanism \cite{tsai2019multimodal} or an inter-modal temporal position prediction task \cite{yu2023speech} to capture cross-modal alignment information. The latter addresses data annotation's time-consuming and labor-intensive nature through semi-supervised methods. Liang \cite{liang2020semi} designed a cross-modality distribution matching task to enhance the consistency of emotional representation, while Lian \cite{lian2022smin} proposed a Semi-supervised Multi-modal Interaction Network (SMIN) to learn multimodal interactive and contextual information. 
\begin{figure}[tbp]
\centering
\includegraphics[scale=0.75]{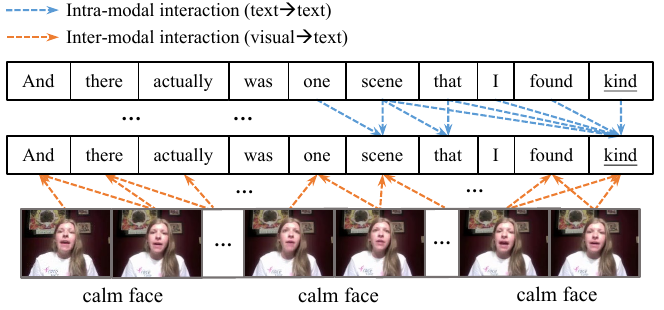}
\caption{The importance of dynamically controlling the intra- and inter-modal interactive information. The arrows denote attention weights, whereas the blue and orange arrows indicate attention weight distributions "between words in language modeling" and "from visual to textual modalities", respectively. For instance, on the semantic similarity level, the words "scene" and "that" refer to the same concept, resulting in higher attention scores between them (arrow: "scene" to "that"). On the task-oriented level, the word "kind" is a key sentiment word and thus has a higher self-attention score (arrow: "kind" to "kind").}
\label{figex}
\end{figure}
Current supervised and semi-supervised learning methods have made significant advancements, but they struggle to incorporate sentiment-related features from unlabeled data or properly separate irrelevant knowledge specific to different modalities. Figure~\ref{figex} illustrates how crucial it is to maintain a balanced proportion of information passing between intra- and inter-modal interactions. On one side, the speaker's emotion can be inferred from key sentiment words such as ``kind''. Nevertheless, the transfer of information between visual and textual sequences in an inter-modal interactive manner can lead to confusion, as conventional attention may not distinguish between different words (due to the absence of clear positive emotional cues in image sequences) and overlook important emotional expressions. Conversely, when a speaker's comments, gestures, and intonation all convey a uniform emotional tone, inter-modal interaction becomes crucial. It can be used as additional information for interaction within the same mode. Therefore, we suggest a new framework that adjusts the proportion of intra- and inter-modal interaction based on the assumption that independent learning and dynamic selection of information are essential. This framework also utilizes self-training to gain knowledge from unlabeled data. Our contributions can be summarized as follows:
\begin{itemize}
\item We present a new network called Semi-IIN that combines two unique masked attention mechanisms to capture meaningful interactions among image sequences, audio frames, and text tokens.
\item We use a self-training method that creates dependable pseudo-labels by a top-k confidence filtering strategy, allowing for model improvements through retraining and the extraction of emotion-related features from data without labels. Under the semi-supervised learning setting, Semi-IIN achieves improved performance.  
\item Experimental results on two public datasets show that Semi-IIN performs better than other current methods. To better understand how effective our approach is, we carry out thorough ablation experiments.
\end{itemize}

\section{RELATED WORK}
\subsection{Semi-supervised sentiment analysis}
Supervised learning methods are commonly used for sentiment analysis, but their effectiveness is limited by the lack of labeled data for training. Researchers are trying to address this challenge by incorporating semi-supervised learning techniques to decrease the need for labeled data and enhance overall performance. To generate reliable pseudo-samples, researchers are committed to incorporating consistency-based pseudo-label strategy to identify misleading instances \cite{yuan2024multimodal}, or establishing a specific threshold for prediction confidence in categories with clear and dependable characteristics \cite{cheng2023semi}. Another direction is utilizing autoencoders \cite{lian2022smin, zhang2020multi} to extract emotion-salient representations from additional unlabeled data.
To narrow the heterogeneous gap between different modalities, Hu \cite{hu2020semi} designed a Semi-supervised Multimodal Learning Network, which correlates different modalities by capturing the multimodal data's intrinsic structure and discriminative correlation. Liang \cite{liang2020semi} proposed a semi-supervised learning method based on cross-modal distribution matching. Parthasarathy \cite{parthasarathy2020semi} employed semi-supervised ladder networks that incorporated skip connections between the encoder and decoder to extract emotion-relevant features. 
\subsection{\textbf{Multimodal interaction learning}}
Previous research has focused on creating fusion strategies to capture interactive connections. Existing methods can be categorized into utterance-level and token-level interaction learning. To facilitate learning at the level of utterance interactions, single-mode representations are initially encoded individually and then combined by applying constraints \cite{hazarika2020misa, yu2021learning}, separating \cite{tsai2018learning}, analyzing correlations \cite{sun2020learning}, or capturing relationships \cite{zadeh2017tensor, liu2018efficient} to enable single-mode, dual-mode, or triple-mode interactions. Recently, Han \cite{han2021improving} introduced information theory to maximize the mutual information between unimodal and multimodal fusion results, while Yang \cite{yang2023confede} performed contrastive representation learning and contrastive feature decomposition to enhance the representation of multimodal information. Nevertheless, these approaches fail to account for the fact that emotions can vary throughout different points in the video as unimodal representations are averaged along the time axis to capture intricate and evolving emotional signals. Additionally, they face either high computational complexity or the introduction of extra hyperparameters. For token-level interaction learning, MAG-BERT proposed by Rahman \cite{rahman2020integrating}, incorporates non-verbal token-level information by generating a shift based on visual and acoustic modalities to enhance interaction learning. Nevertheless, this method necessitates coordination among modalities. To address this challenge, Tsai \cite{tsai2019multimodal} proposed a cross-modal attention mechanism to reinforce the target modality with emotional signals from the source modality. Recently, Chen \cite{chen2023inter} proposed an inter-intra modal representation augmentation approach to enhance modal-representation learning ability.  

Our goal is to utilize sentiment information from text, audio, and visual cues at the token level to improve the model's generalization ability with the help of semi-supervised learning. In contrast to previous studies, the processing of interactive information across different senses is carried out independently, filtering out distracting stimuli and regulated by a gate mechanism to maintain the consistency of emotional cues. Moreover, we use a self-training approach to enhance model training.

\section{METHOD}
\begin{figure*}[ht]
\centering
\includegraphics[width=1\textwidth]{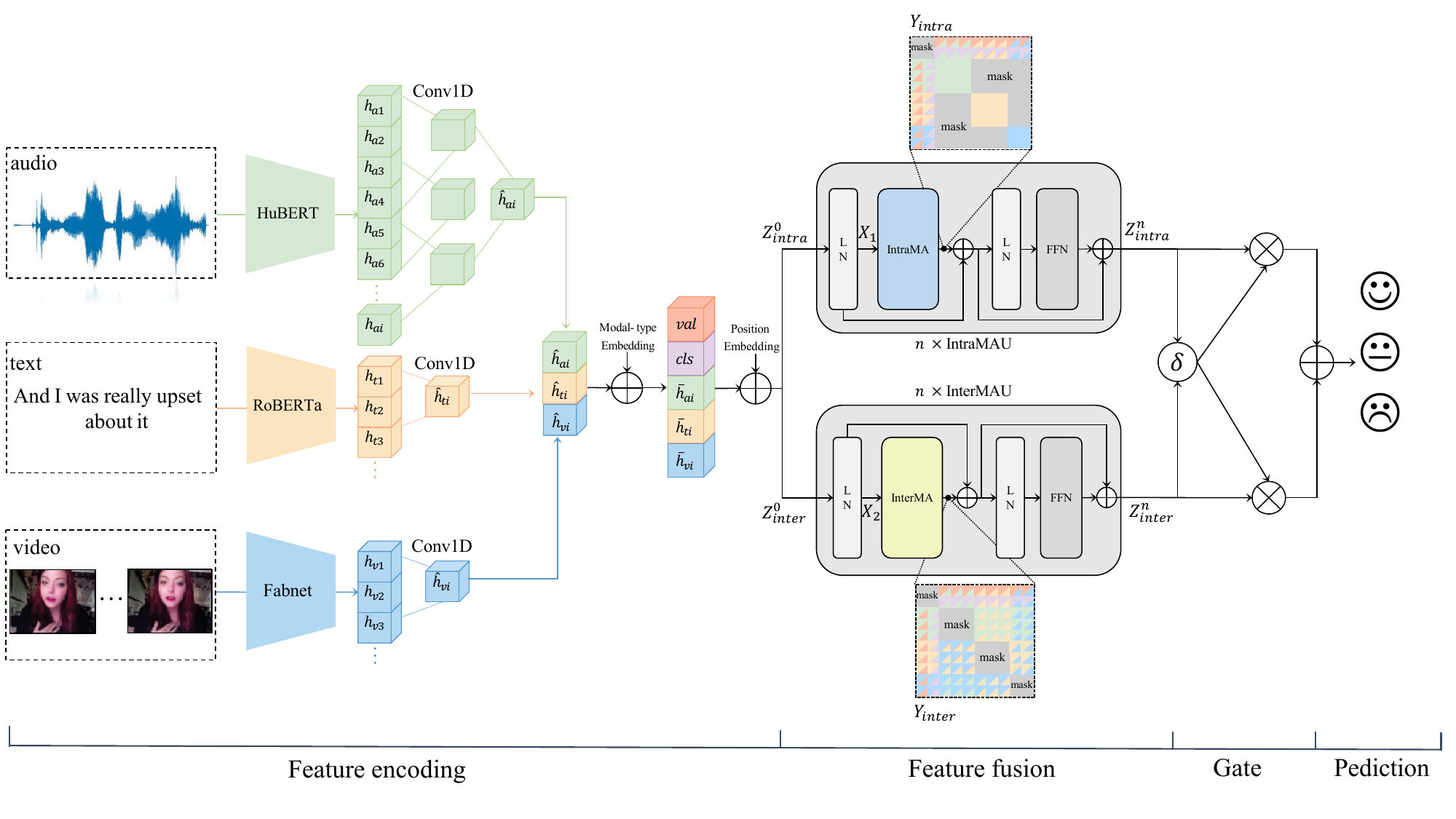}
\caption{The overall architecture of Semi-IIN. Notably, $Z^{0}_{inter}$ and $Z^{0}_{intra}$ are the same as $Z$ in equation \eqref{Z}.}
\label{figart}
\end{figure*}
\begin{figure}[ht]
\centering
\includegraphics[width=0.47\textwidth]{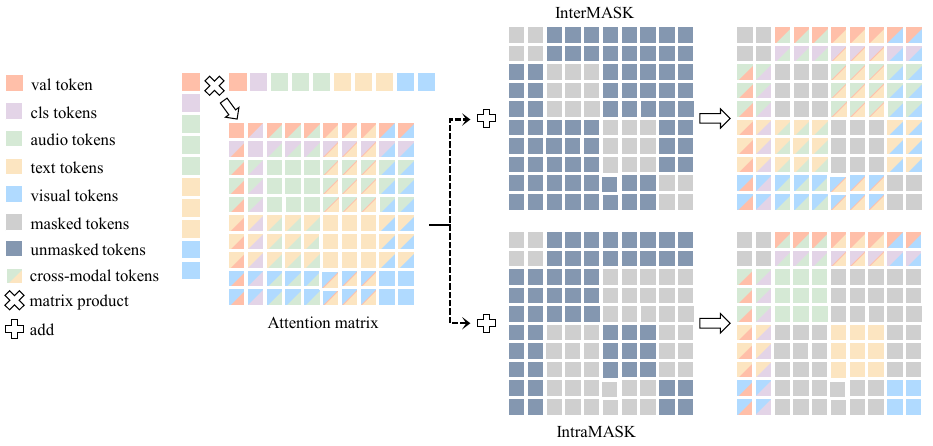}
\caption{Implementation of InterMA(top) and IntraMA(bottom).}
\label{figma}
\end{figure}
\subsection{Feature encoding}
This section provides a detailed description of our proposed framework, Semi-IIN. As shown in Figure~\ref{figart}, we first use the pre-trained 24-layer  RoBERTa \cite{liu2019roberta} to capture lexical features to obtain a single-modal representation, following the approach of previous studies \cite{lian2022smin, chen2023inter}. To capture visual emotions, we use the pre-trained Fabnet \cite{wiles2018self, chen2023inter} to depict fundamental emotional characteristics. HuBERT \cite{hsu2021hubert} is utilized for extracting the initial vector representations in the context of acoustic modality. They are formulated as:
\begin{equation}
    h_{ti}=RoBERTa(X_{ti};\theta_{t}^{RoBERTa})\in \mathbb{R}_{}^{l_t\times d_t}
\end{equation}
\begin{equation}
    h_{vi}=Fabnet(X_{vi};\theta_{v}^{Fabnet} )\in \mathbb{R}_{}^{l_v\times d_v}
\end{equation}
\begin{equation}
    h_{ai}=HuBERT(X_{ai};\theta_{a}^{HuBERT} )\in \mathbb{R}_{}^{l_a\times d_a}
\end{equation}
where $h_{ti}$, $h_{vi}$, and $h_{ai}$ represent features corresponding to the lexical, visual, and acoustic modalities for the $i$-th sample, respectively. 
Next, a module for local feature extraction is used, which includes an 1D convolutional neural network (Conv1D) with various receptive fields to uncover the emotion-relevant features of each type of data. They are:
\begin{equation}
    \hat{h}_{ti}=Conv1D(h_{ti},k_{t} )\in \mathbb{R}_{}^{l_{t}^{'}\times d_{h}}
\end{equation}
\begin{equation}
    \hat{h}_{vi}=Conv1D(h_{vi},k_{v})\in \mathbb{R}_{}^{l_{v}^{'}\times d_{h}}
\end{equation}
\begin{equation}
    \hat{h}_{ai}=Conv1D(Conv1D(h_{ai},k_{a1}),k_{a2})\in \mathbb{R}_{}^{l_{a}^{'}\times d_{h}}
\end{equation}
where $k_{t}$, $k_{v}$, $k_{a1}$, and  $k_{a2}$ are the convolutional kernel sizes. $d_h$ is the common hidden dimension. Afterward,  $\hat{h}_{ti}$, $\hat{h}_{vi}$, and $\hat{h}_{ai}$ are added with their corresponding modal-type embeddings.
\begin{equation}
    \overline{h}_{ti}=\hat{h}_{ti} + t_{}^{type}
\end{equation}
\begin{equation}
    \overline{h}_{vi}=\hat{h}_{vi} + v_{}^{type}
\end{equation}
\begin{equation}
    \overline{h}_{ai}=\hat{h}_{ai} + a_{}^{type}
\end{equation}

Next, we combine the hidden representations at the token level from three different modalities using two weight vectors ($cls$ and $val$), along with introducing positional encoding ($PE$), to create the multimodal input sequence $Z$. That is:
\begin{equation}
\label{Z}
    Z=[val;cls;\overline{h}_{ai}^{1};...;\overline{h}_{ai}^{L};...;\overline{h}_{ti}^{1};...;\overline{h}_{ti}^{N};\overline{h}_{vi}^{1};...;\overline{h}_{vi}^{M}]+PE
\end{equation}
where $Z \in \mathbb{R}_{}^{T \times d_h}$. $L$, $N$, and $M$ represent the length of input feature sequences of corresponding modalities and the total sequence $T=L+N+M+2$.

\subsection{Feature fusion}
As shown in Figure \ref{figma}, two distinct Masked Attention(MA), Intra-modal Masked Attention(IntraMA) and Inter-modal Masked Attention(InterMA) are designed to mask unrelated relations between modalities. The former focuses on exploiting interactive information within each modality, while the latter enables the cross-modal exchange of emotional cues. Specifically, we design two different attention masks: Intra-modal MASK(IntraMASK) and Inter-modal MASK(InterMASK):
\begin{equation}
\centering
\text{IntraMASK}_{ij}=
\begin{cases}
   0,    & \text{ if } i,j \in \text{Intra}_{pos} \\
   -\infty, & \text{ if }    i,j  \notin \text{Intra}_{pos}

\end{cases}
\end{equation}
\begin{equation}
\centering
\text{InterMASK}_{ij}=
\begin{cases}
   0,    & \text{ if } i,j \in \text{Inter}_{pos} \\
   -\infty, & \text{ if }    i,j  \notin \text{Inter}_{pos}

\end{cases}
\end{equation}
where IntraMASK $\in \mathbb{R}_{}^{T \times T}$ and InterMASK $\in \mathbb{R}_{}^{T \times T}$. $\text{Intra}_{pos}$ and $\text{Inter}_{pos}$ are two pre-defined matrices designed to separate the tokens of the interaction position from the masked ones. After that, IntraMA is achieved by adding the IntraMASK with the conventional global attention \cite{vaswani2017attention}, which facilitates the extraction of the key emotional clues in each modality. It is mathematically expressed as: 
\begin{equation}
\centering
\begin{aligned}
    Y_{intra}&=\text{IntraMA}(X_1) \\
     &=softmax(\frac{QK_{}^{T}}{\sqrt{d_{k}}}+\text{IntraMASK})V \\ 
     &=softmax(\frac{X_1W_{Q}W_{K}^{T}{X_{1}^T}} {\sqrt{d_{k}}}+\text{IntraMASK})X_1W_{V} 
\end{aligned}
\end{equation}
where input $X_1\in\mathbb{R}_{}^{T\times d_{h}},W_{Q}\in\mathbb{R}_{}^{d_{h}\times d_{k}},W_{K}\in\mathbb{R}_{}^{d_{h}\times d_{k}}$, and $W_{V}\in\mathbb{R}_{}^{d_h\times d_{v}}$. 
Similarly to the IntraMA, InterMA is achieved as:
\begin{equation}
\centering
\begin{aligned}
    Y_{inter}&=\text{InterMA}(X_2) \\
     &=softmax(\frac{QK_{}^{T}}{\sqrt{d_{k}}}+\text{InterMASK})V \\ 
     &=softmax(\frac{X_2W_{Q}W_{K}^{T}{X_2}^T} {\sqrt{d_{k}}}+\text{InterMASK})X_2W_{V} 
\end{aligned}
\end{equation}

Subsequently, as shown in Figure~\ref{figart}, by replacing the global attention with the proposed IntraMA and InterMA, the Intra-modal Masked Attention Unit(IntraMAU) and the Inter-modal Masked Attention Unit(InterMAU) are constructed. Specifically, for the $l$-th layer input $Z^l_{intra} \in \mathbb{R}_{}^{T \times d_h}$, the output of $l$-th layer of the IntraMAU can be calculated as:  
\begin{equation}
\begin{aligned}
 \hat{Z}_{intra}^{l}&=\text{IntraMA}(\text{LN}(Z_{intra}^{l-1}))+\text{LN}(Z_{intra}^{l-1}) \\
    Z_{intra}^{l}&=\text{FFN}(\text{LN}(\hat{Z}_{intra}^{}))+\text{LN}(\hat{Z}_{intra}^{l})
\end{aligned}
\end{equation}
where $l \in [1,n]$, FFN and LN represent the feed-forward network with ReLU as the activation function and the layer normalization, respectively. The InterMAU consists of the same modules as the IntraMAU except for the IntraMA being replaced with the InterMA.
\subsection{Gate mechanism}
Two special tokens, $val$, and $cls$, serve as fusion features, aggregating information from all tokens except for the special ones. Thus, the first element of $Z_{intra}^{n}$ and $Z_{inter}^{n}$, along with the second position of $Z_{intra}^{n}$ and $Z_{inter}^{n}$ are treated as the final fusion feature. They are processed via the following dynamic gate mechanism \cite{lv2021progressive}:
\begin{equation}
\begin{aligned}
\centering
    G_{v} &= sigmoid(Z_{intra}^{n}[0] \cdot W_{1}+Z_{inter}^{n}[0] \cdot W_{2}+b_{1}) \\
    Z_{v} &= G_{v}\odot Z_{inter}^{n}[0] + (1-G_{v}) \odot Z_{intra}^{n}[0] \\ 
    G_{e} &= sigmoid(Z_{intra}^{n}[1] \cdot W_{1}^{'}+Z_{inter}^{n}[1] \cdot W_{2}^{'}+b_{1}^{'}) \\
    Z_{e} &= G_{e}\odot Z_{inter}^{n}[1] + (1-G_{e}) \odot Z_{intra}^{n}[1]
\end{aligned}
\end{equation}
where $W_{1}$, $W_{2}$, $W_{1}^{'}$ and $W_{2}^{'}$ all $\in \mathbb{R}_{}^{d_h \times d_{h}}, b_{1}$ and $b_{1}^{'} \in \mathbb{R}_{}^{d_h}$. The passed proportions between modality-specific and modality-complimentary knowledge are further dynamically determined through the gating approach, facilitating meaningful modality interaction learning. The predictions of sentiment intensity $v$ and emotional category $e$ are derived from the filtered feature $Z_{v}$ and $Z_{e}$. They are
\begin{equation}
    v=\text{FC}_{v}(\text{MLP}(Z_{v})) \in \mathbb{R}_{}^{d_{1}^{}}
\end{equation}
\begin{equation}
    e=\text{FC}_{e}(\text{MLP}(Z_{e})) \in \mathbb{R}_{}^{d_{2}^{}}
\end{equation}
\begin{algorithm}[t]
\caption{Self-training}
\label{alg}
\textbf{Input}: Labeled dataset($L_d$) and unlabeled dataset($U_d$). Model  $\phi$\\
\textbf{Output}: Final model $\phi'$
\begin{algorithmic}[1] 
\WHILE{current epoch $<$ total epoch}
\FOR{$sample_i$, $v_i$, $e_i$ in $L_d$}
\STATE $\hat{v}_i$, $\hat{e}_i$ = $\phi$($sample_i$)
\STATE Using equation \eqref{eqloss1}, \eqref{eqloss2} and \eqref{eqlabloss} to caculate loss and update model $\phi$'s parameters.
\ENDFOR
\ENDWHILE
\STATE Initialize model $\phi'$ using $\phi$. Generate predictions based on $U_d$ and $\phi'$. Constructing dataset $U'_d$ by selecting the top-k highest confidence ones in each category
\WHILE{current epoch $<$ total epoch}
\FOR{$sample_i$, $v_i$, $e_i$ in $L_d \cup U'_d$}
\STATE $\hat{v}_i$, $\hat{e}_i$ = $\phi'$($sample_i$)
\STATE Using equation \eqref{eqloss1}, \eqref{eqloss2}, \eqref{eqloss3} and \eqref{equlabloss} to caculate loss and update model $\phi'$'s parameters.
\ENDFOR
\ENDWHILE
\STATE \textbf{return} model $\phi'$
\end{algorithmic}
\end{algorithm}
\begin{table*}[htbp]
\small
\centering
\renewcommand\arraystretch{1.5} 
\setlength{\tabcolsep}{3.7pt} 
\begin{tabular}{c|c|cccccccccc}
\hline
         &           & \multicolumn{5}{c}{MOSI}                                                & \multicolumn{5}{c}{MOSEI}                          \\ \cline{3-12} 
Methods  & Embedding & MAE$\downarrow$& Corr$\uparrow$& Acc-2$\uparrow$& F1$\uparrow$& \multicolumn{1}{c|}{Acc-7$\uparrow$} & MAE$\downarrow$& Corr$\uparrow$& Acc-2$\uparrow$& F1$\uparrow$& Acc-7$\uparrow$\\ \hline
$\text{LMF}^\dag$ &           Glove
& 0.917 & 0.695 & -/82.50     & -/82.40     & \multicolumn{1}{c|}{33.20
}  & 0.623 & 0.700 & -/82.00     & -/82.10     & 48.00  
\\
$\text{TFN}^\dag$ &           Glove
&

0.901 & 0.698& -/80.80     & -/80.70     & \multicolumn{1}{c|}{34.90
}  & 
0.593 & 0.677 & -/82.50     & -/82.10     & 50.20  
\\
$\text{MFM}^\ddag$&           Glove
& 0.877 & 0.706 & -/81.7      & -/81.6      & \multicolumn{1}{c|}{35.40
}  & 0.568 & 0.703 & -/84.40     & -84.30      & 51.30  \\
$\text{Mult}^\ddag$&           Glove
& 

0.861 & 0.711 & 81.50/84.10 & 80.60/83.90 & \multicolumn{1}{c|}{-
}  & 
0.580 & 0.713 & 82.50/84.23 & 82.67/83.97 & -      \\
$\text{ICCN}^\ddag$&           Bert-base
& 0.862 & 0.714 & -/83.00     & -/83.00     & \multicolumn{1}{c|}{39.00
}      & 0.565 & 0.704 & -/84.20     & -/84.20     & 51.60  \\
$\text{MISA}^\dag$&           Bert-base
& 

0.804 & 0.764 & 80.79/82.10 & 80.77/82.03 & \multicolumn{1}{c|}{-}      & 0.568 & 0.717 & 82.59/84.23 & 82.67/83.97 & -      \\
$\text{Self-MM}^\dag$&           Bert-base
& 0.713 & 0.798 & 84.00/85.98 & \underline{84.42}/85.95& \multicolumn{1}{c|}{-
}      & 0.530 & 0.765 & 82.81/85.17 & 82.53/85.30 & -      \\
$\text{MAG-BERT}^\ddag$&           Bert-base
& 

0.712 & 0.796 & \underline{84.20}/86.10& 84.10/86.00 & \multicolumn{1}{c|}{-
}      & -     & -     & \underline{84.70}/-& \underline{84.50}/-& -      
\\
$\text{MMIM}^\dag$&           Bert-base
& 0.700& 0.800 & 84.14/86.06 & 84.00/85.98 & \multicolumn{1}{c|}{\textbf{46.65}}  & 0.526 & 0.772 & 82.24/85.97 & 82.66/85.94 & 54.24  
\\
$\text{ConFEDE}^\dag$&           Bert-base
& 

0.742 & 0.784 & 84.17/85.52 & 84.13/85.52& \multicolumn{1}{c|}{42.27}  & 0.522 & 0.780 & 81.65/85.82 & 82.17/85.83 & \underline{54.86}\\
$\text{SMIN}^\dag$&           Roberta-large
& -     & -     & -/81.55     & -/81.45     & \multicolumn{1}{c|}{-}      & 
-     & -     & -/86.82     & -/86.81     & -      \\
$\text{TCDN}^\dag$&           Roberta-large& \underline{0.697}& \underline{0.805}& -/\textbf{87.10}& -/\textbf{87.20}& \multicolumn{1}{c|}{-}  & \underline{0.521}& \underline{0.782}& -/\underline{87.50}& -/\underline{87.20}& -\\ \hline
$\text{Ours}^\dag$&           Roberta-large& 

\textbf{0.679}& \textbf{0.822}& \textbf{85.28}/\underline{87.04}& \textbf{85.19}/\underline{87.00}& \multicolumn{1}{c|}{\underline{46.50}}  & 
\textbf{0.497}& \textbf{0.804}& \textbf{84.98/87.70}& \textbf{85.27/87.65}& \textbf{55.89}\\ \hline
\end{tabular}
\caption{Results on CMU-MOSI and CMU-MOSEI dataset. The best performance is highlighted in bold, while the second-best is denoted with an underline. †: unaligned setting. ‡: aligned setting } \label{tab1_2}
\end{table*}
\subsection{Self-training}
We design a self-training strategy to distill emotional knowledge from unlabeled data. The process is shown in algorithm \ref{alg}. Initially, we train the MSA model shown in Figure~\ref{figart} with labeled data and then employ the trained model, referred to as $\phi$, to make predictions using unlabeled data. Following recent progress in semi-supervised learning \cite{chen2023semi}, the top-k confidence method is employed to eliminate unreliable samples. Due to the significantly larger amount of data in the MOSEI dataset in comparison to the MOSI dataset, the model trained on the MOSEI dataset generates predictions with greater confidence and accuracy. As a result, we assigned a value of 40 to k for the MOSI dataset, whereas, for the MOSEI dataset, k is set to the total number of unlabeled instances. Finally, we combine the labeled and unlabeled data(the labeled portion) to retrain the model(weight initialization from $\phi$). Note that only the emotion classification task loss is sent back through the network for pseudo-labeled samples.

\subsection{Loss Function}
Emotional states can be represented either through discrete categories (such as ``sad'' and ``happy'') or dimensional annotations (points in a continuous space). In the MSA, Lian \cite{lian2023mer} and Wang ~\cite{wang2022emotional} highlighted a high correlation between discrete and dimensional annotations. As a result, we classify the data into seven specific emotional categories by determining how close the dimensional labels are to predefined discrete categories. After obtaining discrete labels, we adopt the Mean Squared Error (MSE) $L_{v}$ and Cross-entropy Loss $L_{e}$ as our optimization objectives. We have:
\begin{equation}
\label{eqloss1}
    L_{v}=\frac{1}{N_{l}}{\sum_{i}^{N_{l}}(\hat{v}^{i}-v^{i})^2} 
\end{equation}
\begin{equation}
\label{eqloss2}
    L_{e}=-\frac{1}{N_{l}}{\sum_{i}^{N_{l}}e_{i}log(\hat{e}^{i})}
\end{equation}
\begin{equation}
\label{eqloss3}
    L_{e}^{u}=-\frac{1}{N_{u}}{\sum_{i}^{N_{u}}e_{i}log(\hat{e}^{i})}
\end{equation}
The loss function is defined in the supervised learning process as follows:
\begin{equation}
\label{eqlabloss}
    L_{total}^{p}=\lambda_{1} L_{v}+(1-\lambda_{1})L_{e}
\end{equation}
while for the semi-supervised process, the loss function is as follows:
\begin{equation}
\label{equlabloss}
    L_{total}^{r}=\lambda_{1} L_{v}+(1-\lambda_{1})L_{e}+\lambda_{2}L_{e}^{u}
\end{equation}
where $N_{l}$  and $N_{u}$ are the number of labeled and unlabeled training samples, respectively. $\lambda_{1}$ and $\lambda_{2}$ are two weighting factors. 

\section{Experiment}
\subsection{Dataset}
\textbf{CMU-MOSI} \cite{zadeh2016mosi}, a dataset for human MSA, includes 2,199 video segments taken from 93 videos. Each segment is marked with a sentiment score between -3 and +3 to show the level of sentiment expressed in that portion.\\
\textbf{CMU-MOSEI} \cite{zadeh2018multimodal}, an enhanced version of MOSI, consists of 22,856 video clips. Each segment is annotated with sentiment and emotion.\\
\textbf{AMI} \cite{carletta2005ami} dataset  includes 100 hours of meeting recordings. It provides video recordings of each speaker, voice track, and transcripts of their speeches. We use it as the unlabeled dataset because it does not include any sentiment annotation.

\subsection{Evaluation metrics}
We report Mean Absolute Error (MAE), Pearson correlation (Corr), binary classification accuracy (Acc-2), and F1-score on MOSI and MOSEI datasets. The Acc-2 and F1-score are calculated in negative/non-negative (non-exclude zero) and negative/positive (exclude zero). 

\subsection{Implementation Details }
Semi-IIN is trained with the Adam optimizer, configured with a learning rate of 1e-4. The batch size is 32. The loss weight factors $\lambda_{1}$ and $\lambda_{2}$ are set at 0.6 and 0.2, and the embedding size of transformer encoders is 128. We implement the proposed Semi-IIN on a single NVIDIA A100.
\begin{figure}[t]
\centering
    \includegraphics[width=\linewidth]{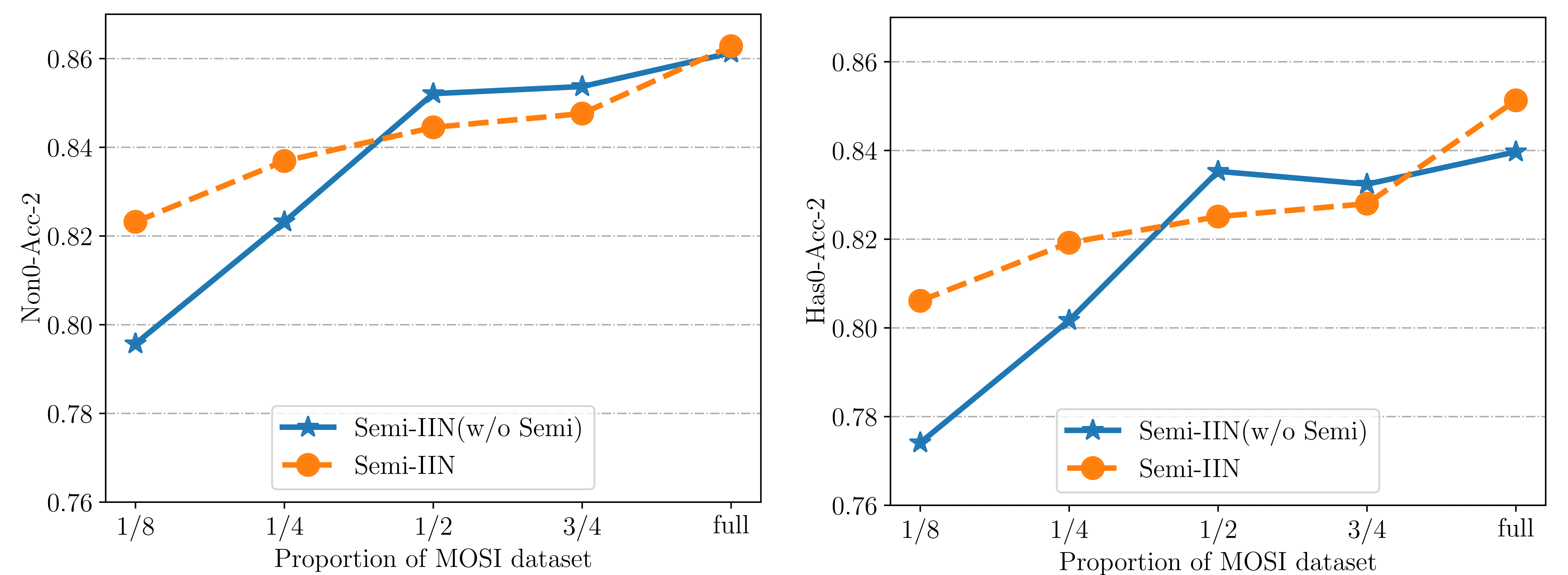}
  \caption{Results under different proportions of labeled samples on MOSI dataset.}
  \label{semi}
  \includegraphics[width=\linewidth]{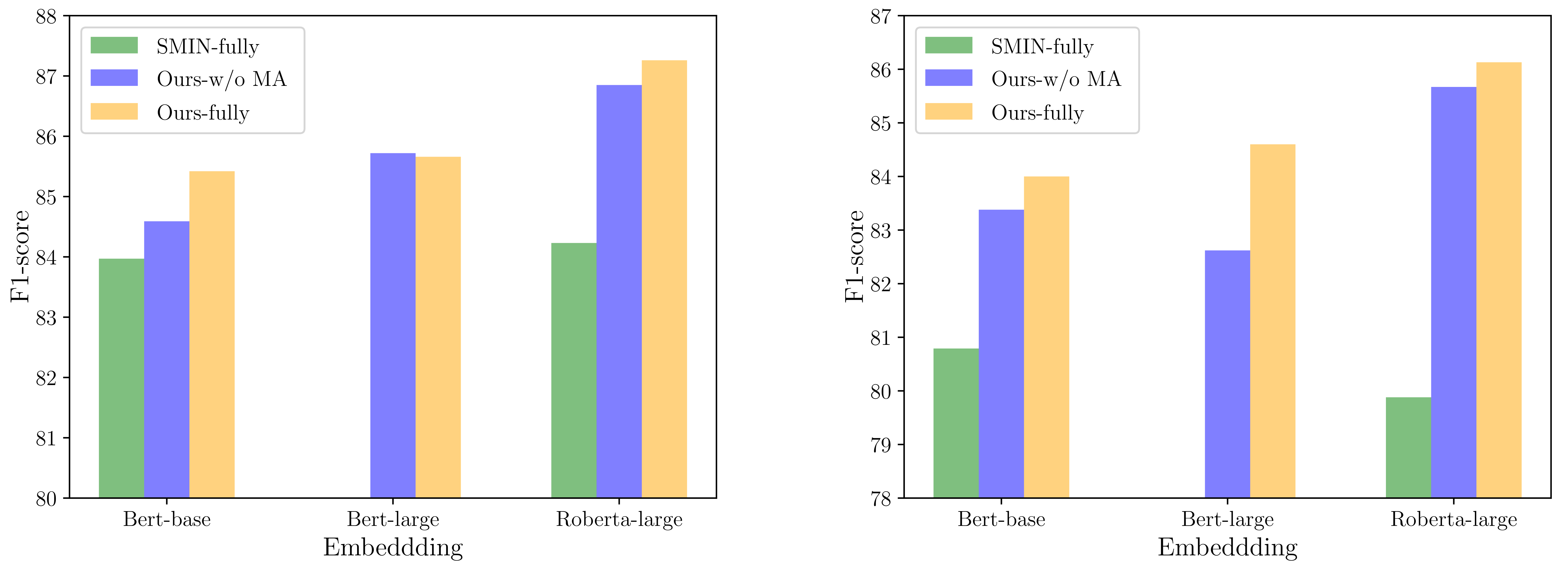}
  \caption{Comparison between different embedding on two datasets. Left: MOSEI dataset. Right: MOSI dataset. SMIN-fully: the previous semi-supervised SOTA method(under fully-supervised training). Ours-w/o MA: Semi-IIN without MA is trained under full supervision. Ours-fully: Semi-IIN with MA is trained under full supervision.}
  \label{embed}
\end{figure}
\subsection{Comparison to State-of-the-art Methods}
Table~\ref{tab1_2} illustrates the results for two datasets. For the unaligned setting (LMF, TFN, MISA, Self-MM, MMIM, ConFEDE, SMIN, and TCDN) and the aligned setting (MFM, ICCN, MulT, and MAG-BERT), our method achieves competitive performance on the MOSI and MOSEI datasets. Specifically, compared to the current SOTA method TCDN which employs the same word embeddings (Roberta-large), Semi-IIN surpasses it by 0.016 MAE and 0.017 Corr on the MOSI dataset, respectively. On the MOSEI dataset, Semi-IIN surpasses TCDN by 0.024 MAE and 0.45\% accuracy. These results demonstrate the superiority of Semi-IIN in MSA. 

\subsection{Ablation study}
Firstly, to verify the effectiveness of the self-training strategy, we conducted experiments with varying ratios of labeled samples in a semi-supervised training scenario. Figure~\ref{semi} illustrates that Semi-IIN still shows progress even with a small number of labeled samples. However, performance decreases when the ratio is adjusted to 50\% or 75\%. The decrease in numbers could be due to the lack of balance in the created fake samples.
\begin{table}[tbp]
\small
\renewcommand\arraystretch{1.5}
 \setlength{\tabcolsep}{1pt}
\begin{tabular}{c|c|c|c|c|c|c|c}
\hline
Method& MA& Semi & Params & MAE$\downarrow$& Corr$\uparrow$& Acc-2$\uparrow$& F1$\uparrow$\\ \hline
Baseline &                                                                -&      -& 1.3M   & 0.509 & 0.793 & 83.97/86.85& 84.30/86.8\\ \hline
Semi-IIN    & \checkmark                                                              &      &        1.6M& 0.499 & 0.800 & \textbf{85.04}/87.26& 85.22/87.14 \\
         &                                                                & \checkmark    & 1.3M   & 0.507 & 0.792 & 84.54/86.74 & 84.8/86.64  \\
         & \checkmark                                                              & \checkmark    & 1.6M   & \textbf{0.497}& \textbf{0.804}& 84.98/\textbf{87.70}& \textbf{85.27/87.65}\\ \hline
\end{tabular}
\caption{Comparison of the overall result of Semi-IIN with different settings. Baseline employs conventional global attention. MA: IntraMA and InterMA. Semi: Semi-supervised learning} \label{tabab2}
\end{table}
Additionally, experiments are carried out to confirm that the increase in performance is not a result of improving word embeddings. Since Lian \cite{lian2022smin} only shares findings from fully supervised learning with different embedding setups, we conducted training for Semi-IIN in a fully supervised manner to maintain a fair comparison. As illustrated in Figure~\ref{embed}, Semi-IIN-fully consistently outperforms Semi-IIN-fully (without MA) across different embedding configurations on both the MOSI and MOSEI datasets. This result confirms the efficacy of the masked attention strategy. Furthermore, compared to SMIN-fully, Semi-IIN-fully demonstrates superior performance, validating its suitability and scalability. 

Furthermore, various ablation experiments are carried out under different conditions to showcase the effectiveness of the suggested MA, along with the semi-supervised learning approach. The results are presented in Table \ref{tabab2}. Compared with the baseline model, despite introducing a few parameters, Semi-IIN(only MA) achieves nearly 1\% accuracy improvement and lower MAE. Semi-IIN(with Semi) also results in a 0.5\% increase in accuracy. The best outcome is achieved by combining Semi and MA. The findings above show that separating interactions into two branches from both intra- and inter-modal perspectives helps to better utilize consistent emotional signals across modalities. Additionally, using additional unlabeled data slightly improves the training of the model.

We also utilize various fusion methods to confirm the importance of dynamically determining the proportions of intra- and inter-modal information, as shown in Table \ref{tabab3}. The findings show that the gating fusion method is more effective than other fusion techniques, highlighting the importance of choosing between modality-specific and modality-common information.

\begin{table}[tbp]
\small
\renewcommand\arraystretch{1.3}
\begin{tabular}{c|cccc}
\hline
Fusion mode& MAE$\downarrow$& Corr$\uparrow$& Acc-2$\uparrow$& F1$\uparrow$\\ \hline
dot        &       0.506&       0.794&             84.07/87.01&             84.37/86.93\\ \cline{1-1}
add       &       0.512&       0.794&             83.45/86.85&             83.80/86.78\\ \cline{1-1}
concat      & 0.506  & 0.794      & 83.64/86.65 & 84.03/86.64          \\ \cline{1-1}
gate        & \textbf{0.499}& \textbf{0.800}& \textbf{85.04/87.26}& \textbf{85.22/87.14}\\ \hline
\end{tabular}
\caption{The impact of different fusion modes(Based on Semi-IIN(only MA))} \label{tabab3}
\end{table}
\begin{figure*}[t]
\centering
\includegraphics[width=0.7\linewidth]{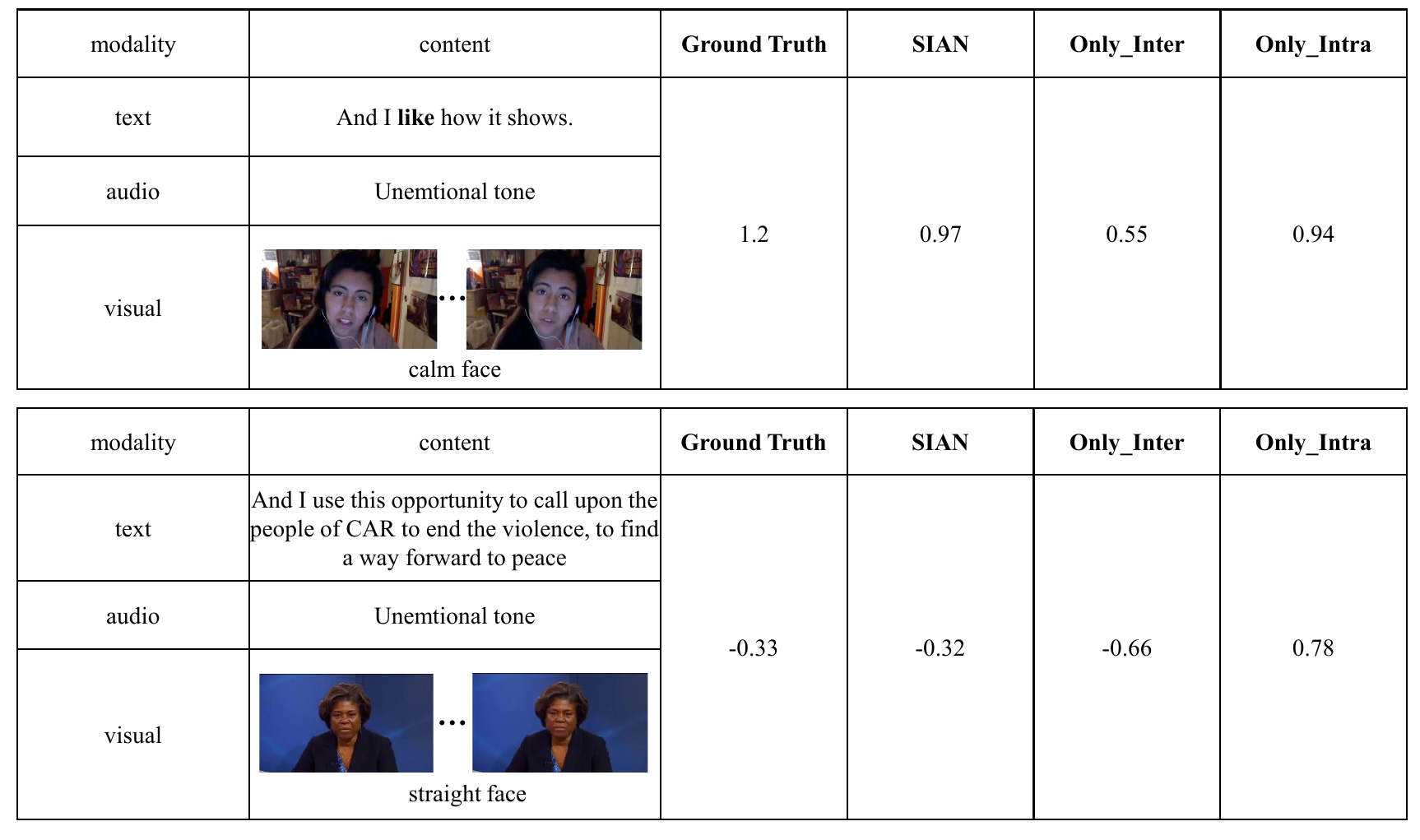}
\caption{Case study for the Semi-IIN. The ``Only Intra'' and the ``Only Inter'' refer to the stacked IntraMAU and InterMAU prediction, respectively.}
\label{vfigex}
\end{figure*}
\begin{figure}[t]
\centering
\includegraphics[width=\linewidth]{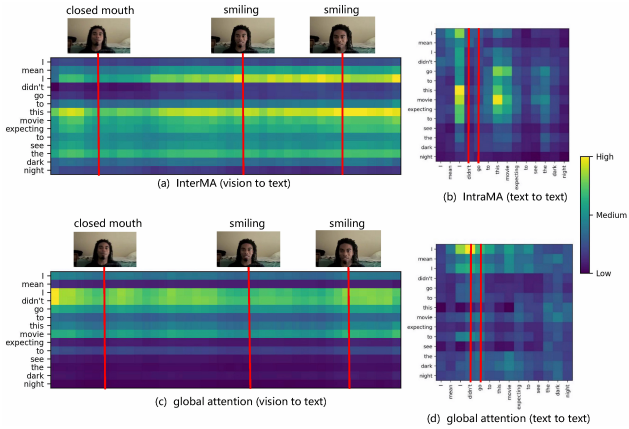}
\caption{Visualization of IntraMA and InterMA mechanisms} 
\label{vfigma}
\end{figure}
\section{Qualitative Analysis}
\subsection{Case study}
In this section, we select two examples to verify the importance of dynamically selecting effective interactions. As shown in Figure~\ref{vfigex}, in the first case, since visual and acoustic modality both contain irrelevant emotional signals such as calm face and unemotional tone, the inter-modal interactive branch is inclined to perceive the speaker's emotions as closer to neutral. In contrast, the intra-modal interactive branch offers a more precise prediction by disregarding the unuseful cross-modal information flow. In the second case, the intra-modal interactive branch is affected by the predominant lexical mode, leading to an inaccurate sentiment evaluation. Conversely, the inter-modal interactive branch fully utilizes visual modality that consists of abundant sentiment cues to reinforce lexical and acoustic modality, leading to accurate emotional polarity. In these two cases, Semi-IIN remains unaffected by irrelevant interactive noise and achieves accurate results overall by effectively exploring interactions.

\subsection{Visualization of IntraMA and InterMA}
Figure~\ref{vfigma} illustrates the visualization results of InterMA and IntraMA. It is noteworthy that the speaker's emotion in this video is positive. Figure~\ref{vfigma}(a) and Figure~\ref{vfigma}(c) show that the InterMA pays more attention to important image frames with a lot of emotional content, instead of focusing on unnecessary frames, e.g., neutral facial expressions, unlike traditional global attention. Moreover, Figure~\ref{vfigma}(b) and Figure~\ref{vfigma}(d) illustrate how the IntraMA mechanism mitigates the impact of emotion-unrelated words like ``didn’t go'', which may lead to incorrect affective polarity, by assigning less attention compared to conventional global attention. We think that IntraMA and InterMA are effective because they can use specific and complementary knowledge from different modalities to filter out unnecessary information.


\section{Conclusion and Future Work}
This paper introduces a new MSA framework, called Semi-IIN, aimed at reducing both intra- and inter-modal noise at a detailed level. Semi-IIN, along with the IntraMA and InterMA mechanisms, successfully captures important interactive information within and between modalities, making it easier to extract consistent emotional cues from multimodal data. In addition, our model decreases the need for extensive human annotations by including semi-supervised learning. The effectiveness of Semi-IIN is demonstrated through experimental results on two benchmark datasets, CMU-MOSI and CMU-MOSEI. Our proposal outperforms previous approaches, setting the new SOTA result for MSA. Our future directions mainly lie in designing semi-supervised intra-inter modal interaction learning networks for multilingual multimodal sentiment analysis, e.g., Spanish, French, and German, and enhancing interpretability. 

\section{Acknowledgements}
This work was supported in part by the National Natural Science Foundation of China under Grant 62202174, in part by the Basic and Applied Basic Research Foundation of Guangzhou under Grant 2023A04J1674, in part by The Taihu Lake Innocation Fund for the School of Future Technology of South China University of Technology under Grant 2024B105611004, and in part by Guangdong Science and Technology Department  Grant 2024A1313010012.

\bibliography{Semi_IIN}

\onecolumn
\section{Supplementary Materials}
\subsection{Comparison with recent method}
From Table~\ref{tabs1} and \ref{tabs2}, we observe that ours performs comparable performance with the SOTA MMML in terms of various metrics on MOSI. Note that MMML exploits fine-tuning modality feature extractor to improve performance, resulting in an increase in trainable parameters. Our training parameters are significantly lower than MMML, at about 0.4\%. On the MOSI and  MOSEI dataset, our training time is only 90 seconds and 20minutes, respectively. While MMML is 60 minutes and 170 minutes, respectively.
\begin{table}[htbp]
\center
\renewcommand\arraystretch{1.8}
\begin{tabular}{cccccccccc}
\toprule[2pt]
MOSI     & context & finetune & \begin{tabular}[c]{@{}c@{}}training\\ params\end{tabular} & \begin{tabular}[c]{@{}c@{}}training\\ time\end{tabular} & Acc2$\uparrow$        & F1$\uparrow$          & Acc7$\uparrow$  & MAE$\downarrow$   & Corr$\uparrow$  \\ \hline
MMML     &   \XSolidBrush      &   \Checkmark       & 411.92M                                                   & 60min                                                   & 85.28/87.50 & 85.24/87.54 & 47.48 & 0.629 & 0.846 \\ 
Semi-IIN &   \XSolidBrush      &    \XSolidBrush      & 1.6M                                                      & 90s                                                     & 85.28/87.04 & 85.19/87.00 & 46.50 & 0.679 & 0.822 \\ \toprule[2pt]
\end{tabular}
\caption{Result on MOSI dataset}
\label{tabs1}
\end{table}
\begin{table}[htbp]
\center
\renewcommand\arraystretch{1.8}
\begin{tabular}{cccccccccc}
\toprule[2pt]
MOSEI     & context & finetune & \begin{tabular}[c]{@{}c@{}}training\\ params\end{tabular} & \begin{tabular}[c]{@{}c@{}}training\\ time\end{tabular} & Acc2$\uparrow$        & F1$\uparrow$          & Acc7$\uparrow$  & MAE$\downarrow$   & Corr$\uparrow$  \\ \hline
MMML     &   \XSolidBrush      &  \Checkmark        & 411.92M                                                   & 170min                                                   & 81.28/86.52 & 81.96/86.60 & 54.67 & 0.510 & 0.795 \\ 
Semi-IIN &    \XSolidBrush     & \XSolidBrush         & 1.6M                                                      & 20min                                                     & 84.98/87.70 & 85.27/87.65 & 55.89 & 0.497 & 0.804 \\ \toprule[2pt]
\end{tabular}
\caption{Result on MOSEI dataset}
\label{tabs2}
\end{table}

\subsection{Label acquisition details}

The emotional categories in MOSEI and MOSI datasets are turned into
seven categories by mapping the dimensional sentiment intensity to discrete emotional categories, as shown in the following Table. 
\begin{longtable}[]{@{}ll@{}}
\toprule[2pt]\noalign{}
emotional category & dimensional label \\
\midrule\noalign{}
\endhead
\endlastfoot
0 & {[}-3, -2.5) \\
1 & {[}-2.5, -1.5) \\
2 & {[}-1.5, -0.5) \\
3 & {[}-0.5, 0.5) \\
4 & {[}0.5, 1.5) \\
5 & {[}1.5, 2.5) \\
6 & {[}2.5, 3{]} \\
\bottomrule[2pt]\noalign{}
\caption{The mapping relation between emotion category and dimensional label}\\
\end{longtable}

The mapping aims to ensure the optimization direction of classification and regression loss consistent in both fully supervised and semi-supervised learning stages. During self-training, it introduces more reliable emotional labels rather than dimensional labels and composes "softer" constraints for model optimization. 


\end{document}